\documentclass[twoside,11pt]{article}

\usepackage{times}
\usepackage{listings}
\usepackage{lscape}
\usepackage[abbrvbib,preprint]{jmlr2e}
\usepackage{natbib,subfigure}

\usepackage{lipsum,booktabs}
\usepackage{amsmath,mathrsfs,amssymb,amsfonts,bm}
\usepackage{bbm, dsfont}
\usepackage{paralist}
\usepackage{rotating}
\usepackage{pdflscape}
\usepackage{hyperref,url,dsfont,nicefrac}
\hypersetup{
    colorlinks,
    breaklinks,
    linkcolor = blue,
    citecolor = blue,
    urlcolor  = blue,
}
\allowdisplaybreaks[2]
\usepackage{appendix}
\usepackage{multirow,makecell}
\usepackage{graphicx,color} 
\usepackage{caption2}
\usepackage{standalone}     
\usepackage{preview}
\usepackage{xcolor}
\usepackage{setspace}
\usepackage{lmodern}
\usepackage{algorithmic,algorithm}
\renewcommand{\algorithmicrequire}{ \textbf{Input:}}
\renewcommand{\algorithmicensure}{ \textbf{Output:}}

\renewcommand{\tilde}{\widetilde}
\renewcommand{\hat}{\widehat}

\def \X {\mathcal{X}}

\def \g {\mathbf{g}}

\def \p {\mathbf{p}}

\def \x {\mathbf{x}}
\def \y {\mathbf{y}}
\def \z {\mathbf{z}}

\usepackage{inconsolata} 

\def \epsilon {\varepsilon}

\usepackage{mathtools}

\def \p {\boldsymbol{p}}

\let\proof\relax
  
\usepackage{amsthm}
\let\proof\relax

\newenvironment{proof}{\par\noindent{\textbf{Proof}\ }}{\hfill\BlackBox\\[2mm]}

\newtheorem{myProp}{Proposition}

\theoremstyle{definition}

\newtheorem{myExample}{Example}
\newtheorem{myRemark}{Remark}

\usepackage{hyperref}
\usepackage{url}
\usepackage{listings}
\usepackage{pifont}
\usepackage{multirow}
\usepackage{wrapfig}
\usepackage{ragged2e}
\usepackage[most]{tcolorbox}
\definecolor{darkerlogocolor}{RGB}{20, 0, 145}
\newtcolorbox{ttcolorbox}[1][]{colframe=darkerlogocolor, colback=darkerlogocolor!4!white, title=#1}
\def \algshort {UFPL}

\usepackage{geometry}
\usepackage{lastpage}
\jmlrheading{1}{2024}{1-\pageref{LastPage}}{4/00}{10/00}{Zhang et al}{Zhen-Yu Zhang, Jiandong Zhang, Huaxiu Yao, Gang Niu, and Masashi Sugiyama.}

\ShortHeadings{Unsupervised Prompt Learning with Few-shot Examples for Answering Objective Questions}{Zhang, Zhang, Yao, Niu, Sugiyama}
\firstpageno{1}

\begin{document}

\title{Unsupervised Prompt Learning with Few-shot Examples\\ for Answering Objective Questions}

\author{\name Zhen-Yu Zhang \email zhen-yu.zhang@riken.jp \\
     \addr Center for Advanced Intelligence Project, RIKEN \AND
     \name Jiandong Zhang \email zhang.jiando@northeastern.edu \\
     \addr Northeastern University \AND
     \name Huaxiu Yao \email huaxiu@cs.unc.edu \\
     \addr UNC-Chapel Hill \AND 
     \name Gang Niu \email gang.niu.ml@gmail.com \\
     \addr Center for Advanced Intelligence Project, RIKEN
     \AND
     \name Masashi Sugiyama \email sugi@k.u-tokyo.ac.jp\\
     \addr Center for Advanced Intelligence Project, RIKEN\\
     Graduate School of Frontier Sciences, The University of Tokyo}

\editor{}

\maketitle

\begin{abstract}
\emph{Large language models}~(LLMs) have been highly successful on diverse tasks, while some applications require specialized models to meet stricter accuracy or latency needs; here we focus on objective question answering, a widely encountered and important real-world setting. 
Most existing methods require parameter retraining or human supervision, both entailing high computational and data collection burdens. 
To handle these challenges, a direct approach is to generate ``high-confidence'' data from downstream tasks and use them for prompt learning or in-context learning to refine pseudo-supervision. 
We consider combining both approaches for better performance, but naively learning the prompt first and selecting pseudo supervised examples only at inference creates a mismatch between training and use.
In this paper, we propose \emph{unsupervised few-shot prompt learning}~{(\algshort{})}, which jointly learns the prompt and refines the pseudo-supervision. 
The learning objective \emph{aligns prompt training with usage} by requiring the learned prompt to produce consistent answers when pseudo-supervised data are used as in-context examples. 
We optimize the prompt by using textual feedback to propose prompt updates, and iteratively refine the prompt and the pseudo supervision. 
In a binary classification model, we give a sufficient condition for one prompt update with pseudo-labeled demonstrations to improve expected accuracy across the dataset. 
Empirical results on diverse benchmarks and a real world molecule optimization task show its effectiveness.
\end{abstract}


\section{Introduction}
\label{sec:intro}

Large language models have shown impressive performance on various real-world tasks~\citep{brown2020language,achiam2023gpt}. 
While LLMs are versatile, strict accuracy, latency, or safety targets can motivate \emph{specialization of general purpose LLMs}~\citep{shin2020autoprompt,ouyang2022training}. 
We study \emph{objective question answering}, an important real-world setting where each example has a single verifiable answer. 
It covers many practical tasks that can benefit from specialization, including clinical question answering~\citep{singhal2025toward}, instruction following~\citep{wei2021finetuned}, and so on.

Existing methods for LLM specialization mainly involve fine-tuning, prompt optimization, and in-context learning. 
Parameter efficient fine-tuning, such as LoRA~\citep{hu2021lora}, updates a small subset of parameters to specialize the LLM. 
By contrast, prompt learning~\citep{sun2022black,zhou2022large} and in-context learning~\citep{brown2020language,liu2022makes} adapt behavior by learning input prompts or providing few-shot examples, leaving the base LLM unchanged. 
Although these approaches share the same goal, fine-tuning creates extra model variants that increase storage overhead; for streaming or rapidly changing tasks, repeatedly loading and switching variants adds latency and operational complexity. 
In such cases, retaining a general purpose LLM with task appropriate prompts and data-specific few-shot examples can be preferable~\citep{vu2022spot}.

To facilitate efficient specialization of general purpose LLMs, we adopt prompt learning and in-context learning, \emph{optimizing only the input} to improve downstream performance. 
These methods typically require human supervision to induce prompts or to supply few-shot examples; however, human feedback is costly and time-consuming to collect. 
Fortunately, in many real-world tasks, LLMs can annotate text datasets with quality that matches or exceeds human annotators, motivating the use of unlabeled data with LLMs. 
For example, ChatGPT has been shown to outperform crowd workers on several text-annotation tasks~\citep{gilardi2023chatgpt}, and more broadly LLMs perform strongly on natural language classification benchmarks~\citep{chong2022detecting}. 
Motivated by these observations, we study an unsupervised specialization setting that uses LLM generated pseudo-supervision. 
Our primary goal is to improve predictions on the given unlabeled dataset. 
The resulting pseudo-labeled data can subsequently be used to fine-tune a model for prediction on new data.

It is non-trivial to combine prompt learning and in-context learning to produce refined pseudo-supervision for specialization in an unsupervised setting. 
A naive approach first identifies ``high-confidence'' pseudo-supervised data (e.g., via \emph{chain-of-thought}~{(CoT)} reasoning)~\citep{wei2022chain}, and then either optimizes the prompt using these data~\citep{diao2022black,sun2022black} and uses them as few shot examples for in context prediction~\citep{wan2023universal,guo2024human,lihuman}. 
However, this decoupled design is problematic: it uses few-shot examples only during inference rather than during prompt learning, creating a mismatch between how the prompt is learned and how it is used at inference. 

\begin{figure*}[t!]
    \centering
    \includegraphics[clip, trim=44bp 144bp 18bp 138bp, width=0.98\linewidth]{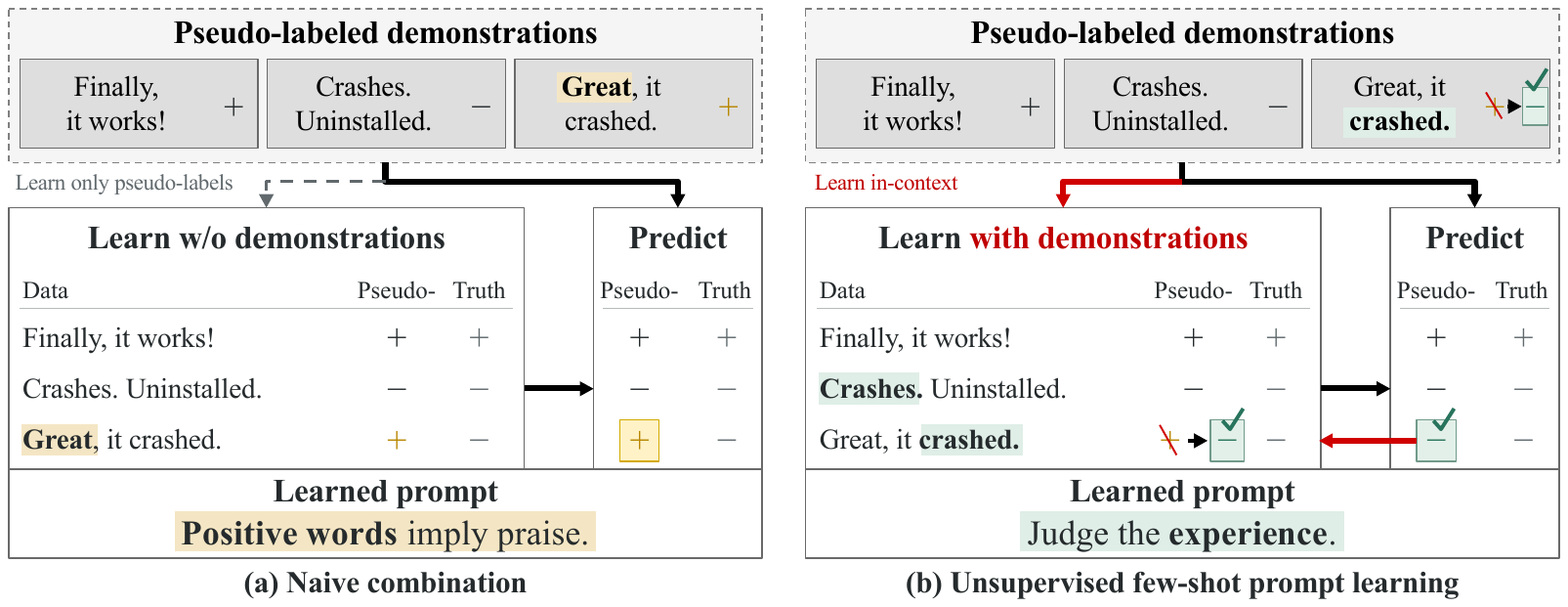}
    \caption{\textbf{Pipeline comparison of \algshort{} and naive combination.} In this example, ``Great, it crashed'' is incorrectly assigned a positive pseudo-label based on the word ``Great''. Learning with pseudo-labeled demonstrations helps the prompt recognize the negative experience shared with ``Crashes. Uninstalled.'' and refine the incorrect pseudo-label. In this way, \algshort{} learns the prompt with pseudo-labeled demonstrations and iteratively refines predictions across the dataset, ensuring consistency between prompt training and usage.}
    \label{fig:main_idea}
    \vspace{-1mm}
\end{figure*}

In this paper, we propose \emph{unsupervised few-shot prompt learning}~(\algshort{}), which learns the prompt with pseudo-labeled demonstrations and iteratively refines predictions across the dataset, ensuring consistency between prompt training and usage, as illustrated in Figure~\ref{fig:main_idea}.
Specifically, we identify ``high-confidence'' pseudo-labeled data and, using these data, align prompt training with usage by requiring the learned prompt to produce consistent answers when pseudo-supervised data from the downstream task are used as in-context examples.
We use TextGrad~\citep{yuksekgonul2024textgrad} to update the prompt using textual feedback, an approach applicable to both open-source and black-box models. 
Theoretical analysis shows that, in the simplified setting of binary classification, \algshort{} can improve expected accuracy over the initial pseudo-labels by learning an optimized prompt with pseudo-labeled demonstrations.
We evaluate \algshort{} with other contenders on several benchmark datasets and a real-world molecule optimization task. 
Experimental results show that \algshort{} produces high-quality refined generations without human supervision, regardless of model scale.

\section{Related Work}
\label{sec:related}

\textbf{Prompt Learning.} Recent advances in prompt learning have developed more systematic methods for prompt design and optimization. Pioneering work such as BBT learns prompts using derivative-free optimization techniques like evolutionary algorithms~\citep{sun2022black}, while BDPL employs policy gradient algorithms to optimize the prompt~\citep{diao2022black}. Gradient-based prompt learning methods have also been proposed, including ProTeGi~\citep{pryzant2023automatic}, TextGrad~\citep{yuksekgonul2024textgrad}, and GREATER~\citep{dasgreater}. In the context of objective QA tasks, these methods typically still require human supervision to optimize the prompt. To learn prompts without labeled data, the Prompt Duel Optimizer~\citep{wu2026llm} asks an LLM judge to compare the outputs of candidate prompts in pairs, while Strategy-Induct~\citep{chen2026strategy} derives a task instruction from reasoning strategies generated for unlabeled questions.

At the level of individual examples, Chain-of-Thought prompting enables LLMs to decompose a hard problem into intermediate steps and solve them sequentially~\citep{wei2022chain}. Subsequent work improves reliability by sampling multiple reasoning paths and selecting the most consistent answer~\citep{wangself}, or explores multiple reasoning branches in a tree-like structure for complex problem solving~\citep{yao2023tree}. These reasoning methods operate on each example independently, leaving predictions on the rest of the downstream data unused.

\textbf{In-context Learning.} In-context learning conditions a frozen LLM on task descriptions and few-shot examples at inference time, enabling rapid adaptation without parameter updates~\citep{brown2020language}. Subsequent analyses investigated what makes ICL effective, highlighting the roles of label space cues, surface form overlap, and example quality~\citep{min2022rethinking, liu2022makes}. A parallel line studied practical levers for ICL, including calibrated decoding to mitigate majority label bias~\citep{zhao2021calibrate} and ordering strategies that reduce sensitivity to example permutations~\citep{lu2022fantastically}. Our work leverages these insights but targets unsupervised specialization: we learn prompts with pseudo-supervised few-shot examples in context so that prompt training is aligned with prompt use at inference.

Recent seminal works have explored using in-context prompting for LLM specialization without retraining model parameters~\citep{wan2023better,wan2023universal,lihuman}. These methods first identify ``high-confidence'' pseudo-supervised data using carefully designed scoring functions, and then leverage the selected data as in-context examples to guide final predictions. PIAST~\citep{batorski2026piast} selects generated demonstrations by estimating their utility, while IterPSD~\citep{gu2026many} iteratively refines pseudo-labels for use in demonstrations. \citet{gadetsky2025large} refined each prediction using other examples paired with their pseudo-labels from the previous round. The proposed \algshort{} uses pseudo-labeled demonstrations during prompt learning and then applies the revised prompt to refine predictions across the dataset.

\textbf{Self-supervised fine-tuning.} In an unsupervised setting, self-supervised fine-tuning methods first use the LLM to generate pseudo-supervised data for a downstream task and then adapt the LLM via parameter-efficient fine-tuning. For instance, LMSI employs CoT prompting~\citep{wei2022chain} to generate high-quality labels for unlabeled datasets, which are then used to optimize the model~\citep{huang2023large}. SELF-ALIGN~\citep{sun2024principle} uses human-written principles and in-context demonstrations to generate responses, which are then used to fine-tune the LLM. A related line of work focuses on output refinement. LLMRefine employs a fine-grained feedback model to identify defects in outputs and guide iterative refinements, optimizing performance during inference without additional training~\citep{xu2024llmrefine}. ISARA is an improved self-refinement method that requires neither human-crafted instructions nor labeled rewards~\citep{guo2024human}.


\def \LLMgen {\textnormal{LM}}
\def \LLMgrad {\textnormal{LM}_{\textnormal{grad}}}
\def \LLMupdate {\textnormal{LM}_{\textnormal{update}}}

\section{Our Approach}
\label{sec:method}

In this section, we introduce the notation, describe the \algshort{} algorithm in detail, and analyze when it improves expected accuracy in a simplified binary classification setting.

\subsection{Notations}
In this part, we introduce the notation. Let $\x_l \in \X$ be the $l$-th example in the unsupervised dataset of size $n$, where $\X$ is the textual space. We denote the prompt by $\z \in \X$ and the initial prompt by $\z_0$. We define the generation function as $\LLMgen(\cdot, \cdot, \cdot): (\x, \z, D) \mapsto \y$, where $\x$ is the input, $\y \in \X$ is the answer in textual space, $\z$ is the prompt, and $D = \{(\x_k, \hat{y}_k)\}_{k=1}^K$ is a set of $K$ pseudo-supervised examples drawn from the downstream task. We denote inference by $\LLMgen(\x, \z, D)$. When $D=\emptyset$, the model predicts without demonstrations. For instance, $\LLMgen(\x, \z_0, \emptyset)$ predicts the answer using the initial prompt alone.

Following TextGrad~\citep{yuksekgonul2024textgrad}, the optimizer compares the model's responses with the labels and produces a textual critique $\p$. A second instruction converts this critique into editing guidance $\g$ for the current prompt:
\begin{equation}
    \label{eqn:grad}
    \g = \LLMgrad(\p).
\end{equation}
The update operator applies this guidance to produce a revised prompt:
\begin{equation}
    \label{eqn:next-prompt}
    \z_{\mathrm{rev}}=\LLMupdate(\z,\g).
\end{equation}

\begin{figure*}[t!]
\centering
\includegraphics[width=\linewidth,trim={42bp 197bp 23bp 206bp},clip]{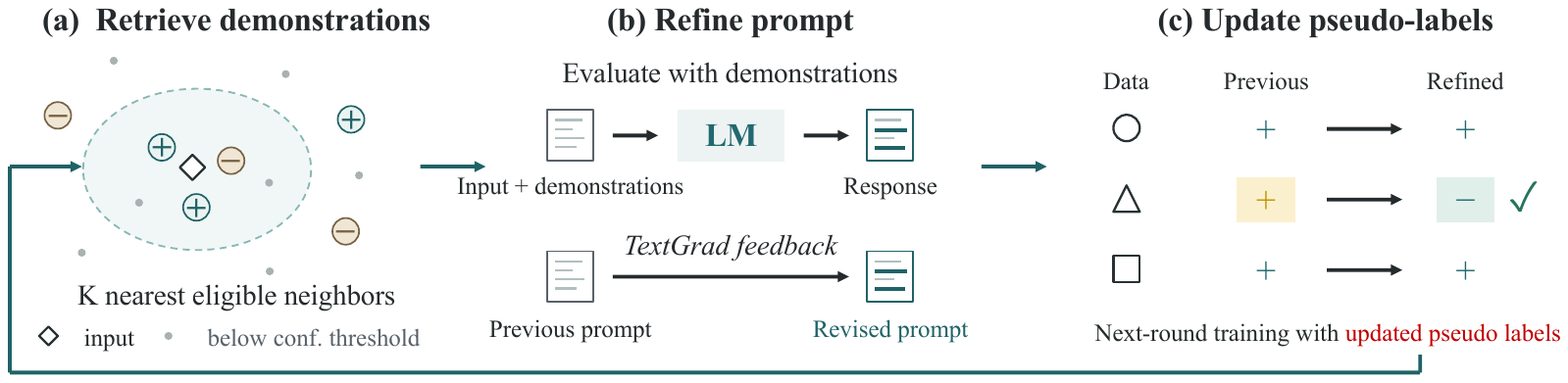}
\caption{\textbf{An illustration of the \algshort{} algorithm.} Each round uses stored pseudo-labels to construct demonstrations, learns a prompt with those demonstrations in context, and applies the revised prompt to refine predictions across the dataset.}
\label{fig:illustration}
\end{figure*}

\subsection{Unsupervised Few-shot Prompt Learning}

In this part, we present the proposed \algshort{} algorithm, which jointly optimizes the prompt and refines the pseudo-supervision for the downstream task in an \emph{iterative} manner.

As in prior work~\citep{wan2023universal,guo2024human,lihuman}, we first consider a straightforward approach that first learns a prompt from pseudo-labeled data and then uses it with demonstrations at inference. Specifically, we initialize the prompt as $\z^{(0)}=\z_0$. For each $\x_l$, we sample $m$ reasoning paths without demonstrations and extract their answers $y_{l_1},\ldots,y_{l_m}$~\citep{huang2023large,wangself}. The answer with the most votes is the initial pseudo-label $\hat y_l^{(0)}$, with confidence
\begin{equation}
    \label{eqn:confidence}
    c_l^{(0)}=\frac{1}{m}\sum_{j=1}^m\mathbbm{1}[y_{l_j}=\hat y_l^{(0)}].
\end{equation}
Let $H_0=\{l:c_l^{(0)}\geq\gamma\}$ be the initial high-confidence set for a threshold $\gamma\in[0,1]$. Write $\ell(y,y')=\mathbbm{1}[y\ne y']$ for disagreement between parsed answers. A baseline that learns a prompt without demonstrations uses the objective
\begin{equation}
    \label{eqn:direct}
    \min_{\z}\sum_{l\in H_0}\ell\bigl(\LLMgen(\x_l,\z,\emptyset),\hat y_l^{(0)}\bigr).
\end{equation}
Notice that this objective learns the prompt without the demonstrations used during inference, creating a mismatch between prompt training and usage.

To reduce this mismatch, \algshort{} learns from the pseudo-labels of high-confidence data and optimizes the prompt with demonstrations in context. At round $t$, let $c_l^{(t)}$ be the fraction of all answers collected so far for $\x_l$ that agree with its current pseudo-label $\hat y_l^{(t)}$. We form $H_t=\{l\in\{1,\ldots,n\}:c_l^{(t)}\geq\gamma\}$ from the entire dataset for both prompt training and demonstration selection. Thus, pseudo-labeled data can enter or leave the high-confidence set as their confidence changes.

Given $H_t$, let $S_l^{(t)}$ index the nearest available neighbors of $\x_l$. The objective at round $t$ is
\begin{align}
    \label{eqn:pod}
    D_l^{(t)} &= \{(\x_k,\hat y_k^{(t)}): k\in S_l^{(t)}\},\nonumber\\
    J_t(\z) &= \sum_{l\in B_t}\ell\bigl(\LLMgen(\x_l,\z,D_l^{(t)}),\hat y_l^{(t)}\bigr),
\end{align}
where $V\subset H_0$ is a fixed held-out set and $B_t\subseteq H_t\setminus V$ is the training batch sampled at round $t$. Using \(J_t\) as the training objective, TextGrad updates the prompt using feedback on the model's responses relative to the pseudo-labels.

We use a fixed held-out set $V$ to check prompt updates. We compare the current and revised prompts based on their predictions' agreement with the current pseudo-labels $\hat y_l^{(t)}$ on $V$, keeping these labels and the demonstrations fixed during the comparison. The revised prompt is retained only if this agreement does not decrease. Samples in $V$ are excluded from the prompt-training loss but remain eligible as demonstrations when they belong to $H_t$. Therefore, this check measures pseudo-label consistency within the dataset.

To select these demonstrations, we follow prior work on in-context example selection~\citep{liu2022makes,min2022rethinking} and retrieve $K$ nearest neighbors
\begin{equation}
    \label{eqn:knn}
    S_l^{(t)}=\operatorname{kNN}_K(\x_l,\{k:k\ne l,\ k\in H_t\}).
\end{equation}
Here $\operatorname{kNN}_K$ returns the indices of the nearest available inputs by Euclidean distance between their sentence embeddings~\citep{liu2022makes}. In addition, to mitigate majority label bias in the in-context examples, we adopt the plug-in de-biasing method of~\citet{zhao2021calibrate}.

With the updated prompt $\z^{(t+1)}$ and demonstrations $D_l^{(t)}$, we sample $m$ new answers for each input and pool them with all previous votes to obtain $\hat y_l^{(t+1)}$ by majority voting. We update the confidence scores from the pooled votes and form $H_{t+1}$ from the entire dataset. The updated pseudo-labels of data in $H_{t+1}$ are used as training targets and demonstration labels in the next round.

We illustrate the proposed \algshort{} algorithm in Figure~\ref{fig:illustration} and provide pseudo-code in Algorithm~\ref{alg:all}. Our algorithm proceeds iteratively and can be terminated early depending on the time and cost constraints of the downstream task.

\begin{figure*}[t]
    \centering
\begin{minipage}{0.95\textwidth}
\begin{algorithm}[H]
\caption{Unsupervised Few-shot Prompt Learning (\algshort{})}
    \label{alg:all}
    \begin{algorithmic}[1]
    \REQUIRE Initial prompt $\z^{(0)}=\z_0$, initial vote counts, confidence scores $c_l^{(0)}$ from~\eqref{eqn:confidence}, and the held-out set $V\subset H_0$.
    \FOR{$t=0$ {\bfseries to} $T-1$}
        \STATE Retrieve neighbors using~\eqref{eqn:knn} and construct the demonstrations $D_l^{(t)}$ in~\eqref{eqn:pod} for each $\x_l$.
        \STATE Sample a high-confidence batch $B_t$ from $H_t\setminus V$ and compute $J_t$ in~\eqref{eqn:pod}. 
        \STATE Obtain the revised prompt $\z_{\mathrm{rev}}$ via~\eqref{eqn:grad}--\eqref{eqn:next-prompt}; set $\z^{(t+1)}=\z_{\mathrm{rev}}$, or $\z^{(t+1)}=\z^{(t)}$ if the agreement on $V$ drops.
        \STATE For every $l$, sample $m$ answers with $\z^{(t+1)}$ and $D_l^{(t)}$. Pool these with all previous votes to obtain $\hat y_l^{(t+1)}$, used as training targets and demonstration labels in the next round.
        \STATE Compute $c_l^{(t+1)}$ from all pooled votes and form $H_{t+1}=\{l\in\{1,\ldots,n\}:c_l^{(t+1)}\geq\gamma\}$.
    \ENDFOR
    \STATE Return $\z^{(T)}$ and the answers $\{\hat y_l^{(T)}\}_{l=1}^n$.

\end{algorithmic}
\end{algorithm}
\end{minipage}
\vspace{-2mm}
\end{figure*}

\subsection{Theoretical Analysis}
\label{sec:theorem}

We use a simplified binary classification task to illustrate how \algshort{} learns a better prompt from pseudo-supervision. We analyze when learning the prompt with pseudo-labeled demonstrations yields more accurate predictions across the dataset, in expectation, than the initial pseudo-labels.

Prior work shows that transformers can fit predictors from examples in context~\citep{akyurek2022learning,bai2023transformers}. We model this in-context learning step as empirical risk minimization. Given a prompt, the model fits a classifier that uses the specified attribute and makes the fewest errors on the pseudo-labeled demonstrations. The fitted classifier then predicts the label of the current input.

\begin{myExample}
Consider a fixed dataset $S=\{\x_i\}_{i=1}^n$ with unobserved labels $y_i\in\{-1,1\}$. Each prompt $a$ in a finite family $\mathcal A$ specifies a binary attribute, written $a_i=a(\x_i)$. The attributes may be correlated. For each input $\x_i$, fix a set $N_i$ of $K\ge1$ other indices for use as demonstrations. We assume that one unknown attribute $a_\star\in\mathcal A$ follows the same label mapping within each neighborhood. Specifically, for each $i$, some sign $s_i\in\{-1,1\}$ satisfies $y_j=s_i a_\star(\x_j)$ for all $j\in N_i\cup\{i\}$. The mapping may differ between neighborhoods.

The learner observes pseudo-labels $\tilde y_i$ obtained by independently flipping each true label with probability $1-p$, where $1/2<p<1$. Given prompt $a$, the in-context learner chooses the sign $s$ that minimizes the number of errors $s a_j\ne\tilde y_j$ on $N_i$ and predicts $h_a(i)=s a_i$. Break ties independently for each input and prompt using fair coins independent of the label noise. To learn the shared prompt, draw a batch of $1\le b\le n$ training indices uniformly without replacement, independently of the noise and tie breaks. Select a prompt $\hat a$ whose predictions agree most often with the pseudo-labels on this batch.
\end{myExample}

\begin{myProp}
\label{prop:pseudo-demo-accuracy}
Let $d=|\mathcal A|$, let $q_K$ be the probability that a majority of $K$ independent votes is correct when each is correct with probability $p$, and let $\operatorname{Acc}_S(a)$ be the fraction of the dataset correctly labeled by $h_a$. The expected gain over the initial pseudo-labels satisfies
\[
\mathbb E[\operatorname{Acc}_S(\hat a)]-p
\;\ge\; q_K-p-
\frac{\sqrt{d\bigl(1/b+(K+1)/n\bigr)}}{2(2p-1)}.
\]
Thus the learned prompt improves expected accuracy whenever the right-hand side is positive.
\end{myProp}

\begin{proof}
For $a_\star$, each demonstration votes for the correct local sign $s_i$ with probability $p$, independently of the other votes. The empirical risk minimizer takes their majority, so $\mathbb E[\operatorname{Acc}_S(a_\star)]=q_K$.

Fix a prompt $a$. Write $C_i=\mathbbm{1}[h_a(i)=y_i]$ and $U_i=\mathbbm{1}[h_a(i)=\tilde y_i]$. Since $i\notin N_i$, conditioning on all tie breaks and all label flips except that at $i$ fixes $C_i$ and gives $U_i$ conditional mean $1-p+(2p-1)C_i$. Define the residual
\[
R_i=U_i-\bigl[1-p+(2p-1)C_i\bigr].
\]
It has mean zero and variance $p(1-p)\le1/4$. For $i\ne j$, if $i\notin N_j$, the same conditioning fixes $R_j$, so $\mathbb E[R_iR_j]=0$. Each $N_j$ contains only $K$ indices, and every covariance has magnitude at most $1/4$. Including the diagonal, this gives 
\[
\operatorname{Var}\left(\frac{1}{n}\sum_i R_i\right)
\le \frac{K+1}{4n}.
\]

Let $\widehat U(a)$ be the batch mean of $U_i$. Conditional on the flips and tie breaks, it has mean $n^{-1}\sum_i U_i$ and variance at most $1/(4b)$ by uniform sampling without replacement. Thus the deviation
$D_a=\widehat U(a)-[1-p+(2p-1)\operatorname{Acc}_S(a)]$
satisfies
\[
\mathbb E[D_a^2]
\le \frac{1}{4}\left(\frac{1}{b}+\frac{K+1}{n}\right).
\]

Since $\widehat U(\hat a)\ge\widehat U(a_\star)$ and $\mathbb E[D_{a_\star}]=0$, the expected accuracy loss relative to $a_\star$ is at most $\mathbb E[\max_a D_a]/(2p-1)$. The bound $\mathbb E[\max_a D_a]\le(\sum_a\mathbb E[D_a^2])^{1/2}$ completes the proof.
\end{proof}

\begin{myRemark}
This analysis uses a simplified case to explain the benefit of incorporating pseudo-labeled demonstrations into prompt learning.
For a prompt that uses the predictive attribute, a majority of correctly labeled demonstrations suffices to recover the local label mapping. Each input is excluded from its own demonstrations, and its pseudo-label is correct with probability $p>1/2$. Consequently, among fixed prompts, higher expected accuracy implies higher expected agreement with the training pseudo-labels. The bound shows when the gain from local voting exceeds the cost of selecting a shared prompt from these noisy labels. More generally, learning with pseudo-labeled demonstrations allows \algshort{} to favor prompts that use evidence from neighboring inputs to improve predictions across the unlabeled dataset.
\end{myRemark}


\section{Experiments}
\label{sec:experiments}

In this section, we evaluate \algshort{} with seven contenders across benchmarks. We conduct ablation studies to quantify runtime and cost and to test the contribution of each component. We evaluate \algshort{} on a real-world molecule optimization task.

\subsection{Experimental Setup}
\label{sec:experimental_setup}

In this part, we first introduce the datasets used for comparison, contenders, and implementation details.

\textbf{Datasets.}
We evaluate \algshort{} on benchmarks covering arithmetic reasoning, multiple-choice reasoning and knowledge, commonsense, and natural language understanding: GSM8K~\citep{cobbe2021training}, ASDiv~\citep{miao2020diverse}, MultiArith~\citep{roy2015solving}, AQuA-RAT (AQuA)~\citep{ling2017program}, MMLU~\citep{hendrycksmeasuring} (questions sampled across its 57 subjects), ARC-Challenge (ARC-C)~\citep{clark2018think}, OpenBookQA~\citep{mihaylov2018can}, HellaSwag~\citep{zellers2019hellaswag}, TruthfulQA~\citep{lin2022truthfulqa} (MC1), and SST-2 and RTE from GLUE~\citep{wang2018glue}. All methods operate on unlabeled data, with ground-truth labels used only to evaluate the outputs. We randomly sample 200 questions per task once for evaluation. All reported configurations were selected using separate development questions, disjoint from the evaluation questions, before the evaluation sample was drawn. 

We use $K=5$ demonstrations by default and examine alternative values in Table~\ref{tab:demonstrations}. For RTE, we use the validation split for development and draw evaluation questions from the GLUE training split. Ground-truth labels are never provided to the methods.

\textbf{Contenders.} 
We compare \algshort{} with seven contenders: three baselines that do not use pseudo-supervision (Direct, SC, and Auto-CoT), two in-context learning algorithms, and two prompt learning contenders with pseudo-supervision. Specifically,

\begin{table*}[t]
\centering
\caption{\textbf{Performance comparisons on benchmark datasets.} Accuracy (\%) and mean $\pm$ standard deviation over 3 runs. The best one is in bold. Parenthesized values indicate differences in percentage points relative to the best baseline in each row, computed from the displayed means; gains are red, while drops and ties are black.}
\label{tab:all-tasks-new}
\scalebox{0.72}{
\tabcolsep4pt
\begin{tabular}{l|l|cccccccc}
\hline
\textbf{Model} & \textbf{Dataset} & Direct & Auto-CoT & SC & USP & ICL & SR (BDPL) & SR (TG) & \algshort{} \\
\hline
\multirow{9}{*}{Llama-3.2-1B} & GSM8K & 47.0 $\pm$ 0.0 & 64.2 $\pm$ 1.3 & 64.2 $\pm$ 0.8 & 66.7 $\pm$ 3.2 & 67.7 $\pm$ 2.5 & 45.3 $\pm$ 5.8 & 65.0 $\pm$ 1.8 & \textbf{69.8 $\pm$ 0.8} \textbf{\textcolor{red}{\small{($\uparrow$2.1)}}} \\
 & ASDiv & 70.3 $\pm$ 0.6 & 82.2 $\pm$ 2.8 & 79.5 $\pm$ 0.0 & 81.5 $\pm$ 1.3 & 81.0 $\pm$ 0.9 & 63.7 $\pm$ 4.9 & 79.0 $\pm$ 0.9 & \textbf{83.3 $\pm$ 1.0} \textbf{\textcolor{red}{\small{($\uparrow$1.1)}}} \\
 & MultiArith & 50.0 $\pm$ 0.0 & 80.3 $\pm$ 4.0 & 67.7 $\pm$ 1.3 & 82.8 $\pm$ 5.6 & 81.7 $\pm$ 3.9 & 20.2 $\pm$ 5.5 & 83.2 $\pm$ 10.8 & \textbf{85.7 $\pm$ 1.3} \textbf{\textcolor{red}{\small{($\uparrow$2.5)}}} \\
 & AQuA & 37.5 $\pm$ 0.0 & 53.7 $\pm$ 1.9 & 52.7 $\pm$ 2.3 & 51.5 $\pm$ 1.7 & 52.5 $\pm$ 1.8 & 42.3 $\pm$ 4.6 & 51.2 $\pm$ 2.5 & \textbf{54.0 $\pm$ 0.5} \textbf{\textcolor{red}{\small{($\uparrow$0.3)}}} \\
 & ARC-C & 52.5 $\pm$ 0.0 & 53.7 $\pm$ 1.2 & 53.7 $\pm$ 0.8 & 52.7 $\pm$ 0.3 & 55.3 $\pm$ 1.4 & 53.5 $\pm$ 1.3 & 53.8 $\pm$ 0.8 & \textbf{56.7 $\pm$ 2.4} \textbf{\textcolor{red}{\small{($\uparrow$1.4)}}} \\
 & OpenBookQA & 58.0 $\pm$ 0.0 & 58.5 $\pm$ 1.0 & 59.5 $\pm$ 2.3 & 59.7 $\pm$ 1.0 & \textbf{60.2 $\pm$ 2.5} & 58.0 $\pm$ 0.5 & 58.8 $\pm$ 2.4 & \textbf{60.2 $\pm$ 2.5} \textbf{\textcolor{black}{\small{($\hphantom{\uparrow}$0.0)}}} \\
 & RTE & 50.5 $\pm$ 0.0 & 50.7 $\pm$ 1.0 & 52.5 $\pm$ 0.9 & 51.5 $\pm$ 1.0 & 55.3 $\pm$ 4.3 & 55.5 $\pm$ 1.3 & 52.5 $\pm$ 1.0 & \textbf{57.5 $\pm$ 2.6} \textbf{\textcolor{red}{\small{($\uparrow$2.0)}}} \\
 & MMLU & 43.0 $\pm$ 0.0 & 39.8 $\pm$ 2.5 & 44.3 $\pm$ 1.0 & 40.5 $\pm$ 1.8 & 42.0 $\pm$ 1.8 & 46.2 $\pm$ 1.4 & \textbf{46.3 $\pm$ 1.8} & 45.7 $\pm$ 1.8 \textbf{\textcolor{black}{\small{($\downarrow$0.6)}}} \\
\cline{2-10}
 & Average & 51.1 $\pm$ 0.1 & 60.4 $\pm$ 1.1 & 59.2 $\pm$ 0.2 & 60.9 $\pm$ 1.0 & 62.0 $\pm$ 1.0 & 48.1 $\pm$ 1.3 & 61.2 $\pm$ 2.0 & \textbf{64.1 $\pm$ 0.7} \textbf{\textcolor{red}{\small{($\uparrow$2.1)}}} \\
\hline
\multirow{9}{*}{Qwen3-4B} & GSM8K & 92.5 $\pm$ 0.0 & 93.5 $\pm$ 0.9 & 95.3 $\pm$ 0.3 & 93.7 $\pm$ 0.6 & 94.7 $\pm$ 0.8 & 95.2 $\pm$ 0.6 & 95.0 $\pm$ 0.5 & \textbf{96.0 $\pm$ 0.5} \textbf{\textcolor{red}{\small{($\uparrow$0.7)}}} \\
 & ASDiv & 93.0 $\pm$ 0.0 & 95.7 $\pm$ 0.6 & 95.5 $\pm$ 1.0 & 95.5 $\pm$ 0.5 & 95.3 $\pm$ 0.3 & 95.0 $\pm$ 0.5 & 95.5 $\pm$ 0.5 & \textbf{96.8 $\pm$ 0.5} \textbf{\textcolor{red}{\small{($\uparrow$1.1)}}} \\
 & MultiArith & \textbf{98.0 $\pm$ 0.0} & 97.8 $\pm$ 0.3 & \textbf{98.0 $\pm$ 0.0} & 97.7 $\pm$ 0.3 & \textbf{98.0 $\pm$ 0.0} & \textbf{98.0 $\pm$ 0.0} & 97.5 $\pm$ 0.5 & \textbf{98.0 $\pm$ 0.0} \textbf{\textcolor{black}{\small{($\hphantom{\uparrow}$0.0)}}} \\
 & AQuA & 70.3 $\pm$ 0.8 & 84.3 $\pm$ 1.0 & 82.3 $\pm$ 1.2 & 84.0 $\pm$ 1.5 & 84.3 $\pm$ 0.8 & 81.0 $\pm$ 0.9 & 81.3 $\pm$ 1.0 & \textbf{85.7 $\pm$ 1.3} \textbf{\textcolor{red}{\small{($\uparrow$1.4)}}} \\
 & ARC-C & 93.0 $\pm$ 0.5 & 93.0 $\pm$ 0.9 & 94.3 $\pm$ 0.3 & 94.0 $\pm$ 0.5 & 94.3 $\pm$ 0.8 & 94.2 $\pm$ 0.8 & 94.2 $\pm$ 0.6 & \textbf{95.5 $\pm$ 0.5} \textbf{\textcolor{red}{\small{($\uparrow$1.2)}}} \\
 & OpenBookQA & 89.0 $\pm$ 0.0 & 89.3 $\pm$ 0.6 & 89.3 $\pm$ 0.3 & 90.2 $\pm$ 0.8 & 89.7 $\pm$ 0.3 & 88.2 $\pm$ 1.2 & 88.3 $\pm$ 0.3 & \textbf{90.4 $\pm$ 0.8} \textbf{\textcolor{red}{\small{($\uparrow$0.2)}}} \\
 & RTE & 90.0 $\pm$ 0.0 & 90.2 $\pm$ 0.3 & 90.7 $\pm$ 0.8 & 89.8 $\pm$ 0.6 & 91.3 $\pm$ 0.6 & \textbf{92.3 $\pm$ 0.6} & 89.3 $\pm$ 1.3 & 90.7 $\pm$ 0.3 \textbf{\textcolor{black}{\small{($\downarrow$1.6)}}} \\
 & MMLU & 71.5 $\pm$ 0.0 & 72.3 $\pm$ 2.0 & 70.3 $\pm$ 0.8 & 72.7 $\pm$ 1.5 & 74.5 $\pm$ 1.3 & 75.2 $\pm$ 1.0 & 75.3 $\pm$ 0.8 & \textbf{76.2 $\pm$ 1.2} \textbf{\textcolor{red}{\small{($\uparrow$0.9)}}} \\
\cline{2-10}
 & Average & 87.2 $\pm$ 0.1 & 89.5 $\pm$ 0.3 & 89.5 $\pm$ 0.2 & 89.7 $\pm$ 0.3 & 90.3 $\pm$ 0.3 & 89.9 $\pm$ 0.3 & 89.6 $\pm$ 0.2 & \textbf{91.1 $\pm$ 0.3} \textbf{\textcolor{red}{\small{($\uparrow$0.8)}}} \\
\hline
\end{tabular}
}
\end{table*}

\begin{itemize}
\item \emph{Direct} prompts the LLM with a default prompt to produce answers, and \emph{SC} takes the majority of sampled answers to the same prompt (self-consistency)~\citep{wangself}. \emph{Auto-CoT}~\citep{zhangautomatic} induces intermediate reasoning at inference, encouraging a chain-of-thought before the final answer.
\item Using the same mechanism as \algshort{} to select ``high-confidence'' pseudo supervised examples, we evaluate \emph{ICL}~\citep{liu2022makes}, which uses these examples as few-shot examples to predict the remaining unlabeled data, and \emph{USP}~\citep{wan2023universal}, which scores and selects ``high-confidence'' data and then applies in-context learning for generation.
\item Inspired by the use of self-generated supervision in~\citet{huang2023large}, which performs prompt learning using ``high-confidence'' pseudo-supervised data, we introduce \emph{SR (TG)}, which learns the prompt on pseudo-supervised data with the same textual-gradient optimizer as \algshort{} and keeps an update only if it passes the same held-out check, and \emph{SR (BDPL)}, which learns a discrete prompt of n-grams on the same pseudo-supervised data with the variance-reduced policy gradient of BDPL~\citep{diao2022black}, using the hyperparameters of its official implementation. Unlike \algshort{}, SR (TG) learns the prompt without demonstrations and uses pseudo-labeled demonstrations only at inference. Both methods use the same cumulative voting and demonstration filtering.
\end{itemize}

For fairness, every method except Direct answers from 20 sampled votes, all demonstration-based methods use the same demonstration format, and we apply the plug-in calibration method of~\citet{zhao2021calibrate} to all contenders.

\begin{figure*}[!t] 
\centering
\includegraphics[width=0.97\textwidth]{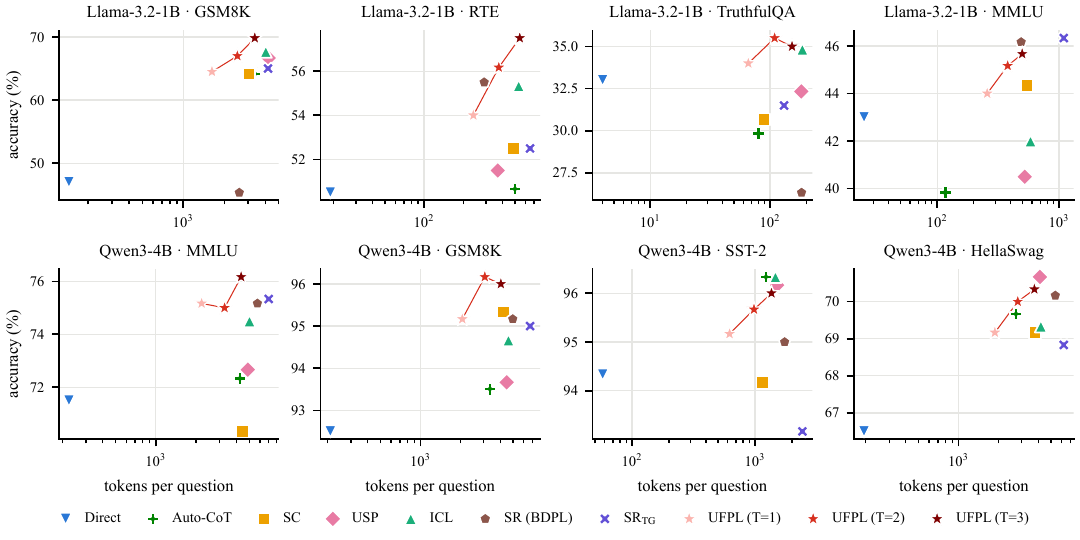}
\caption{ \textbf{Average performance vs.\ cost on Llama-3.2-1B and Qwen3-4B.} Accuracy (\%) against the number of tokens the task model generates per question (log scale), mean over 3 runs. \algshort{} is shown after each round $T=1,2,3$.}
\label{fig:cost}
\end{figure*}

\textbf{Implementation Details.}
For the QA and classification benchmarks, we use Llama-3.2-1B-Instruct~\footnote{\url{https://huggingface.co/meta-llama/Llama-3.2-1B-Instruct}} and Qwen3-4B~\footnote{\url{https://huggingface.co/Qwen/Qwen3-4B}} (with thinking mode off) as task models. Qwen3-14B~\footnote{\url{https://huggingface.co/Qwen/Qwen3-14B}} serves as the optimizer that generates textual critiques and prompt updates for the textual-gradient methods (SR (TG) and \algshort{}). SR (BDPL) uses BDPL as the optimizer. Table~\ref{tab:all-tasks-new} and Fig.~\ref{fig:cost} report means over 3 runs, while the ablation studies report means over 5 runs. We set $T=3$ throughout and use 5 demonstrations for ICL, USP, and \algshort{}. The confidence threshold is fixed at $\gamma=0.65$ for \algshort{} and all competing methods, with $m=5$ sampled answers per example at temperature 0.7.

A prompt update is retained only if it does not lower agreement with the current pseudo-labels on a fixed held-out set of $\min(32,\lfloor|H_0|/3\rfloor)$ data points sampled once from $H_0$. After each round, we recompute confidence as the current pseudo-label's share of all votes collected so far and select $H_{t+1}$ from the entire dataset using $\gamma=0.65$. Both prompt training and demonstration selection use the updated high-confidence set and its current pseudo-labels in the next round. Each question is answered by a majority vote over all answers collected so far, comprising 5 initial answers and 5 per round, for a total of 20 after $T=3$ rounds. The resulting pseudo-labels are used as training targets and demonstration labels in the next round.

Following the official implementation, SR (BDPL) learns a prompt of 20 n-grams, each selected from 50 candidates. It uses 20 sampled prompts per step, a learning rate of $2\times10^{-4}$, and mini-batches of 4, for 10 epochs or at most 200 steps. Each sampled prompt requires two API calls per step, totaling at most 8000 calls. Training uses up to 16 pseudo-labeled questions per label (32 for GSM8K). Section~\ref{sec:prompt-implementation} provides the prompt templates and hyperparameters.

We note that UFPL is not tied to TextGrad: the same training objective can be optimized by other prompt optimization methods, such as evolutionary algorithms~\citep{sun2022black} or variance-reduced policy gradient estimators~\citep{diao2022black}.

\subsection{Performance Comparison on Benchmarks}
In this part, we compare the \algshort{} algorithm with other contenders on benchmark datasets.

\textbf{Performance.}
Table~\ref{tab:all-tasks-new} reports the mean accuracy and standard deviation of \algshort{} and the competing methods. \algshort{} achieves the best or tied-best mean accuracy on most benchmarks, showing strong performance across tasks spanning arithmetic reasoning, knowledge, and natural language understanding. It also achieves the highest average accuracy for both task models, demonstrating that its benefits extend across models with different capabilities.

\algshort{} outperforms Auto-CoT on nearly every benchmark and consistently matches or exceeds Direct, suggesting that learning prompts from unlabeled downstream data can improve predictions beyond chain-of-thought prompting alone. It also achieves higher average accuracy than ICL, USP, SR (BDPL), and SR (TG), highlighting the value of incorporating pseudo-labeled demonstrations into prompt learning rather than using them only at inference. These gains hold for both task models under a common calibration scheme, supporting the benefit of aligning prompt learning with its use at inference.

\textbf{Runtime overhead and cost.}
Our method introduces two sources of overhead: constructing a distance matrix over the unlabeled dataset to select demonstrations at initialization, and refining pseudo-labels and updating the prompt in each round. The initialization cost is amortized across rounds. Each refinement round generates $m$ new answers per question, so the sampling cost grows linearly with $T$.

We compare accuracy and cost over 3 runs with Llama-3.2-1B and Qwen3-4B in Fig.~\ref{fig:cost}. Cost is measured by the number of tokens the task model generates per question, including initial pseudo-labeling where applicable. The optimizer additionally generates about 15 tokens per question for SR (TG) and \algshort{}.

In five of the eight panels, \algshort{} at $T=3$ is more accurate than every contender. It achieves higher accuracy than USP in six panels and generates fewer tokens than USP in seven.

\subsection{Unsupervised Threshold Selection}
Although \algshort{} uses a fixed confidence threshold $\gamma$ by default, it can also select the threshold automatically using the self-consistency scores computed at initialization, without labeled validation data or additional tunable hyperparameters.

For each unlabeled sample $x_i$, let $c_i$ denote its self-consistency score. We sort the scores in descending order,
$c_{(1)} \geq c_{(2)} \geq \cdots \geq c_{(n)}$,
identify the largest adjacent drop,
\[
k^* = \arg\max_{1 \leq k < n} \left(c_{(k)} - c_{(k+1)}\right),
\]
and set
\[
\gamma^* = \max\{0.5, c_{(k^*)}\}.
\]
The threshold is thus determined by the empirical distribution of confidence scores.

Table~\ref{tab:auto_threshold} compares this rule with fixed thresholds of $\gamma=0.65$ and $\gamma=1.0$. The automatically selected threshold is predominantly $\gamma^*=1.0$, so its performance closely follows that of the higher fixed threshold. Compared with the default $\gamma=0.65$, average accuracy decreases by 0.6 percentage points on Llama-3.2-1B and 0.5 on Qwen3-4B. The rule therefore provides an alternative to selecting $\gamma$ using labeled validation data, with a modest reduction in accuracy.

\begin{table}[!htbp] 
\centering
\caption{ \textbf{Effect of automatic threshold.} $\gamma^*$ from the largest drop in sorted vote shares (most frequent value over 5 runs), and \algshort{} accuracy (\%) with $\gamma^*$ and with fixed thresholds.}
\label{tab:auto_threshold}
\resizebox{0.72\linewidth}{!}{%
\begin{tabular}{ll|cccc}
\toprule
\textbf{Model} & \textbf{Dataset} & $\gamma^*$ & Acc.\ ($\gamma^*$) & Acc.\ ($\gamma{=}0.65$) & Acc.\ ($\gamma{=}1.0$) \\
\midrule
Llama-3.2-1B & GSM8K & 1 & 67.9 & 67.8 & 67.9 \\
 & ASDiv & 1 & 81.6 & 83.0 & 81.6 \\
 & MultiArith & 1 & 87.2 & 87.0 & 87.2 \\
 & AQuA & 1 & 53.4 & 54.5 & 53.4 \\
 & ARC-C & 1 & 57.1 & 56.3 & 57.1 \\
 & OpenBookQA & 1 & 60.9 & 60.5 & 60.9 \\
 & RTE & 0.5 & 55.8 & 59.1 & 55.5 \\
 & MMLU & 1 & 45.0 & 45.4 & 45.0 \\
\midrule
Qwen3-4B & GSM8K & 1 & 95.7 & 96.1 & 95.8 \\
 & ASDiv & 1 & 95.6 & 96.8 & 95.6 \\
 & MultiArith & 1 & 98.0 & 98.0 & 98.0 \\
 & AQuA & 1 & 84.6 & 85.7 & 84.6 \\
 & ARC-C & 1 & 94.2 & 95.5 & 94.2 \\
 & OpenBookQA & 1 & 90.3 & 90.5 & 90.3 \\
 & RTE & 1 & 91.4 & 90.9 & 91.4 \\
 & MMLU & 1 & 75.2 & 75.8 & 75.7 \\
\bottomrule
\end{tabular}}
\end{table}

\subsection{Mitigating Error Propagation in Iterative Refinement}

Errors in pseudo-labels can propagate across rounds. \algshort{} mitigates this risk through several mechanisms. Confidence is estimated from multiple sampled answers rather than a single output. Demonstrations are retrieved using distances in the input embedding space, so incorrect pseudo-labels do not affect the neighbor rankings. They are drawn from the current high-confidence set using updated pseudo-labels. We also apply plug-in calibration~\citep{zhao2021calibrate} to mitigate majority label bias in the in-context examples.

We evaluate an additional history-consistency filter, which retains a sample as a demonstration only if its pseudo-label remains unchanged across iterations. This excludes samples with unstable pseudo-labels from the demonstration pool.

Table~\ref{tab:history_consistency} shows that \algshort{} achieves higher average accuracy than the variant with additional history-consistency filtering on both task models, supporting the use of the existing demonstration selection mechanism without this extra restriction.

\begin{table}[!htbp] 
\centering
\caption{ \textbf{Effect of additional history-consistency filtering.} Accuracy (\%), mean $\pm$ standard deviation over 5 runs.}
\label{tab:history_consistency}
\scalebox{0.90}{
\tabcolsep4pt
\begin{tabular}{l|l|cc}
\hline
Model & Dataset & \algshort{} & history-consistency \\
\hline
\multirow{9}{*}{Llama-3.2-1B} & GSM8K & 67.8 $\pm$ 3.3 & 67.2 $\pm$ 1.7 \\
 & ASDiv & 83.0 $\pm$ 1.1 & 82.8 $\pm$ 1.2 \\
 & MultiArith & 87.0 $\pm$ 2.0 & 87.4 $\pm$ 1.9 \\
 & AQuA & 54.5 $\pm$ 1.2 & 54.5 $\pm$ 1.5 \\
 & ARC-C & 56.3 $\pm$ 1.8 & 56.1 $\pm$ 0.9 \\
 & OpenBookQA & 60.5 $\pm$ 1.8 & 60.1 $\pm$ 2.3 \\
 & RTE & 59.1 $\pm$ 3.4 & 55.8 $\pm$ 3.5 \\
 & MMLU & 45.4 $\pm$ 1.3 & 44.8 $\pm$ 1.3 \\
\cline{2-4}
 & Average & 64.2 $\pm$ 0.5 & 63.6 $\pm$ 0.6 \\
\hline
\multirow{9}{*}{Qwen3-4B} & GSM8K & 96.1 $\pm$ 0.4 & 96.1 $\pm$ 0.7 \\
 & ASDiv & 96.8 $\pm$ 0.5 & 96.0 $\pm$ 0.4 \\
 & MultiArith & 98.0 $\pm$ 0.0 & 98.0 $\pm$ 0.0 \\
 & AQuA & 85.7 $\pm$ 1.3 & 84.7 $\pm$ 1.3 \\
 & ARC-C & 95.5 $\pm$ 0.5 & 94.3 $\pm$ 0.6 \\
 & OpenBookQA & 90.5 $\pm$ 1.0 & 90.6 $\pm$ 0.9 \\
 & RTE & 90.9 $\pm$ 0.4 & 91.0 $\pm$ 0.4 \\
 & MMLU & 75.8 $\pm$ 1.0 & 75.8 $\pm$ 1.0 \\
\cline{2-4}
 & Average & 91.2 $\pm$ 0.2 & 90.8 $\pm$ 0.2 \\
\hline
\end{tabular}
}
\end{table}

\subsection{Ablation Studies}
\label{sec:ablation}

We conduct ablation studies to test the contributions of individual components and the effects of key hyperparameters.

\textbf{Effectiveness of each component.}
We evaluate the contribution of each component in \algshort{} through the following ablation variants, using the full method for comparison:
\begin{itemize}
\item \emph{\algshort{} (last round only)}: Predicts each answer using only the last round's votes rather than voting across all rounds.
\item \emph{\algshort{} (w/o update check)}: Retains every prompt update without the held-out agreement check.
\item \emph{\algshort{} (inline label-only demonstrations)}: Presents demonstrations with labels only in a single message rather than as conversation turns with the model's reasoning.
\item \emph{\algshort{} (random sampling)}: Replaces $K$-nearest-neighbor selection with random sampling while maintaining the class balance in the label space.
\item \emph{\algshort{} (single-sample labels)}: Uses pseudo-labels from a single sampled output instead of confidence-based voting.
\item \emph{\algshort{}}: The full method with all components.
\end{itemize}

\begin{figure}[!t]
\centering
\includegraphics[width=0.56\linewidth]{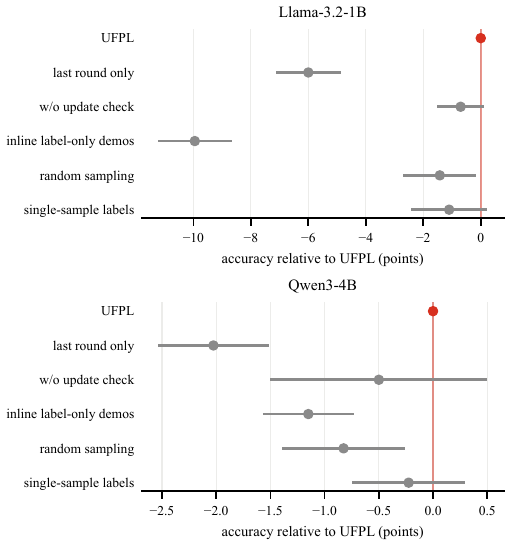}
\caption{\textbf{Ablation study on each component of \algshort{}.} Accuracy of each variant relative to full \algshort{} (points; \algshort{} at 0), averaged over the eight benchmarks, mean $\pm$ standard deviation over 5 runs.}
\label{fig:components}
\end{figure}

\begin{table}[!t] 
\centering
\caption{ \textbf{Number of demonstrations $K$.} \algshort{} accuracy (\%), mean $\pm$ standard deviation over 5 runs.}
\label{tab:demonstrations}
\scalebox{0.90}{
\tabcolsep4pt
\begin{tabular}{l|l|ccc}
\hline
\textbf{Model} & \textbf{Dataset} & $K{=}3$ & $K{=}5$ & $K{=}8$ \\
\hline
\multirow{9}{*}{Llama-3.2-1B} & GSM8K & 67.2 $\pm$ 0.8 & \textbf{67.8 $\pm$ 3.3} & 66.1 $\pm$ 2.3 \\
 & ASDiv & \textbf{84.0 $\pm$ 1.6} & 83.0 $\pm$ 1.1 & 83.9 $\pm$ 1.9 \\
 & MultiArith & \textbf{88.7 $\pm$ 2.0} & 87.0 $\pm$ 2.0 & 87.7 $\pm$ 1.3 \\
 & AQuA & 53.2 $\pm$ 1.5 & \textbf{54.5 $\pm$ 1.2} & 52.7 $\pm$ 2.6 \\
 & ARC-C & \textbf{56.3 $\pm$ 1.2} & \textbf{56.3 $\pm$ 1.8} & 55.2 $\pm$ 0.8 \\
 & OpenBookQA & \textbf{62.0 $\pm$ 1.8} & 60.5 $\pm$ 1.8 & 60.1 $\pm$ 1.9 \\
 & RTE & 56.6 $\pm$ 5.0 & \textbf{59.1 $\pm$ 3.4} & 56.4 $\pm$ 3.1 \\
 & MMLU & 44.0 $\pm$ 1.4 & \textbf{45.4 $\pm$ 1.3} & 43.8 $\pm$ 0.8 \\
\cline{2-5}
 & Average & 64.0 $\pm$ 0.9 & \textbf{64.2 $\pm$ 0.5} & 63.2 $\pm$ 0.4 \\
\hline
\multirow{9}{*}{Qwen3-4B} & GSM8K & 95.6 $\pm$ 0.4 & \textbf{96.1 $\pm$ 0.4} & 95.9 $\pm$ 0.8 \\
 & ASDiv & 96.0 $\pm$ 0.5 & \textbf{96.8 $\pm$ 0.5} & 95.6 $\pm$ 0.7 \\
 & MultiArith & \textbf{98.0 $\pm$ 0.0} & \textbf{98.0 $\pm$ 0.0} & \textbf{98.0 $\pm$ 0.0} \\
 & AQuA & 83.6 $\pm$ 1.3 & \textbf{85.7 $\pm$ 1.3} & 83.8 $\pm$ 0.7 \\
 & ARC-C & 94.0 $\pm$ 0.4 & \textbf{95.5 $\pm$ 0.5} & 94.2 $\pm$ 0.3 \\
 & OpenBookQA & 90.0 $\pm$ 0.7 & \textbf{90.5 $\pm$ 1.0} & 89.8 $\pm$ 1.0 \\
 & RTE & \textbf{91.6 $\pm$ 1.1} & 90.9 $\pm$ 0.4 & 91.1 $\pm$ 0.7 \\
 & MMLU & \textbf{76.6 $\pm$ 0.7} & 75.8 $\pm$ 1.0 & 76.3 $\pm$ 1.1 \\
\cline{2-5}
 & Average & 90.7 $\pm$ 0.3 & \textbf{91.2 $\pm$ 0.2} & 90.6 $\pm$ 0.2 \\
\hline
\end{tabular}
}
\end{table}

Fig.~\ref{fig:components} reports the results with Llama-3.2-1B and Qwen3-4B. Full \algshort{} achieves the highest mean accuracy on both task models, supporting the effectiveness of the complete design. Accumulating votes across rounds and using conversation-turn demonstrations with reasoning provide the largest benefits, highlighting the value of combining predictions from successive rounds and preserving the reasoning information in demonstrations. The update check, nearest-neighbor selection, and confidence-based voting provide additional benefits, helping maintain reliable pseudo-supervision in iterative refinement.

\begin{table}[!t] 
\centering
\caption{ \textbf{Confidence threshold $\gamma$.} Accuracy (\%) mean over 5 runs and the 8 tasks. At initialization, with $m{=}5$ votes, $\gamma{=}0.65$, $0.7$ and $0.8$ keep the same questions ($c \geq 0.8$). Kept: share of questions in the high-confidence set; purity: accuracy of their pseudo-labels.}
\label{tab:gamma-sweep}
\resizebox{0.56\linewidth}{!}{%
\begin{tabular}{ll|cccc}
\toprule
\textbf{Model} & & $\gamma{=}0.4$ & $\gamma{=}0.6$ & $\gamma{=}0.65$ & $\gamma{=}1.0$ \\
\midrule
Llama-3.2-1B & accuracy & 63.5 & 63.7 & 64.2 & 63.6 \\
 & kept & 94\% & 76\% & 49\% & 28\% \\
 & purity & 58.8 & 65.1 & 72.9 & 79.7 \\
\midrule
Qwen3-4B & accuracy & 90.7 & 90.5 & 91.2 & 90.7 \\
 & kept & 97\% & 95\% & 91\% & 85\% \\
 & purity & 91.2 & 91.8 & 93.2 & 94.9 \\
\bottomrule
\end{tabular}}
\end{table}

\begin{table}[!t] 
\centering
\caption{ \textbf{Number of rounds.} \algshort{} accuracy (\%) after $T$ rounds, mean over 5 runs.}
\label{tab:iteration_sensitivity}
\resizebox{0.60\linewidth}{!}{%
\begin{tabular}{ll|ccccc}
\toprule
\textbf{Model} & \textbf{Dataset} & $T{=}1$ & $T{=}2$ & $T{=}3$ & $T{=}4$ & $T{=}5$ \\
\midrule
Llama-3.2-1B & GSM8K & 62.5 & 65.9 & 67.8 & 68.7 & 69.6 \\
 & ASDiv & 80.5 & 82.2 & 83.0 & 82.6 & 83.0 \\
 & MultiArith & 80.7 & 85.7 & 87.0 & 88.5 & 88.6 \\
 & AQuA & 51.2 & 54.1 & 54.5 & 54.3 & 54.4 \\
 & ARC-C & 56.1 & 56.1 & 56.3 & 56.4 & 56.0 \\
 & OpenBookQA & 61.1 & 60.3 & 60.5 & 60.6 & 60.7 \\
 & RTE & 55.0 & 56.3 & 59.1 & 59.0 & 58.2 \\
 & MMLU & 43.9 & 44.8 & 45.4 & 45.1 & 44.3 \\
 & Average & 61.4 & 63.2 & 64.2 & 64.4 & 64.3 \\
\midrule
Qwen3-4B & GSM8K & 95.0 & 95.9 & 96.1 & 96.2 & 96.1 \\
 & ASDiv & 95.4 & 96.0 & 96.8 & 96.1 & 96.0 \\
 & MultiArith & 98.0 & 98.0 & 98.0 & 97.9 & 97.9 \\
 & AQuA & 83.0 & 84.0 & 85.7 & 85.7 & 85.5 \\
 & ARC-C & 94.2 & 94.3 & 95.5 & 94.5 & 94.4 \\
 & OpenBookQA & 90.4 & 90.4 & 90.5 & 90.6 & 90.8 \\
 & RTE & 90.8 & 91.1 & 90.9 & 91.1 & 91.0 \\
 & MMLU & 74.6 & 75.4 & 75.8 & 76.1 & 75.5 \\
 & Average & 90.2 & 90.6 & 91.2 & 91.0 & 90.9 \\
\bottomrule
\end{tabular}}
\end{table}

As shown in Fig.~\ref{fig:components}, these benefits extend across task models with different initial prediction quality. Qwen3-4B produces more accurate initial high-confidence pseudo-labels than Llama-3.2-1B, yet the full method still achieves the highest mean accuracy among the evaluated variants. This suggests that the design of \algshort{} supports refinement both when the initial pseudo-supervision contains substantial errors and when it is already relatively accurate.

\textbf{Number of in-context demonstrations.}
We examine the effect of the number of demonstrations by varying $K$ in nearest-neighbor selection, using the distance metric in~\citet{liu2022makes}. As shown in Table~\ref{tab:demonstrations}, $K=5$ achieves the highest average accuracy for both task models, with $K=3$ performing similarly. These results show that \algshort{} performs effectively with a small number of demonstrations. We therefore adopt $K=5$ as the default, combining strong predictive performance with a compact demonstration context.

\textbf{Confidence threshold.}
We examine the effect of the confidence threshold $\gamma$ for selecting high-confidence pseudo-labeled data, reporting average accuracy over the eight benchmarks in Table~\ref{tab:gamma-sweep}. The default $\gamma=0.65$ achieves the highest average accuracy for both task models, supporting its effectiveness in balancing the quality and size of the selected set. At initialization, the five votes make thresholds of 0.65, 0.7, and 0.8 equivalent for selecting $H_0$. Qwen3-4B also maintains similar accuracy across the evaluated thresholds, indicating robustness to this parameter.

\textbf{Number of Iterations $T$.}
We examine performance over five iterations in Table~\ref{tab:iteration_sensitivity}, complementing the trends in Fig.~\ref{fig:cost}. Accuracy improves through iterative refinement on most tasks across both models, demonstrating the effectiveness of the proposed approach. The results across five rounds show that most of the gains are achieved within the first three iterations, supporting our default choice of $T=3$ for an effective balance between accuracy and computational cost.

\subsection{Molecule Optimization}
\label{sec:molecule}

We also evaluate \algshort{} on a molecular optimization task using DOCKSTRING~\citep{garcia2022dockstring}. Given a molecule represented as a SMILES string, the goal is to propose a candidate with a higher drug-likeness score (QED) and a lower AutoDock Vina docking score~\citep{trott2010autodock}. In this experiment, we use GPT-4o, initialize the prompt with the instruction from TextGrad~\citep{yuksekgonul2024textgrad}, and use five approved drugs as the initial high-confidence examples.

\begin{figure}[t]
\centering
\includegraphics[width=0.45\linewidth]{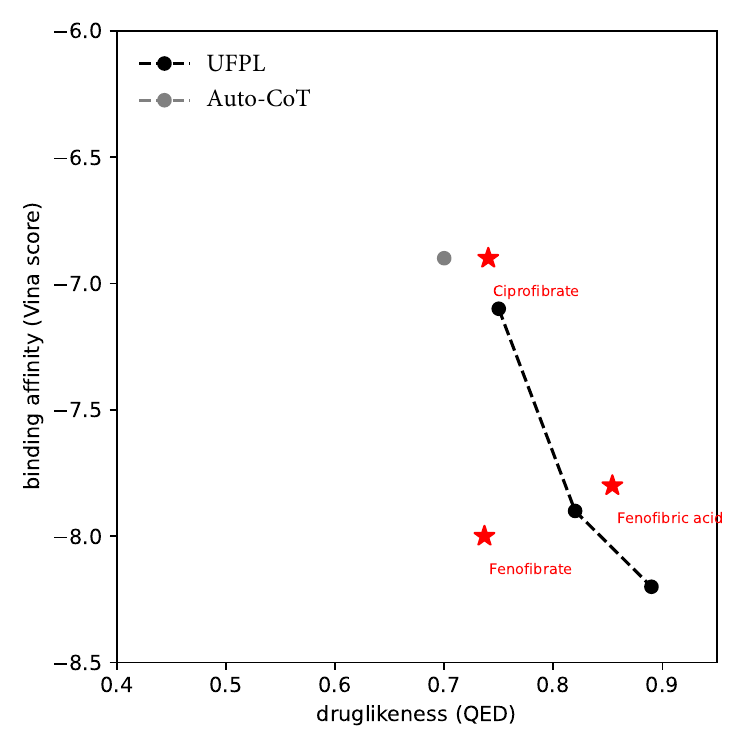}
\caption{QED and Vina scores of the generated molecules. Black circles show the final three \algshort{} iterations, the gray circle shows Auto-CoT, and stars mark the reference compounds.}
\label{fig:molecule}
\end{figure}

Table~\ref{tab:molecule_additional_baselines} compares the five methods. \algshort{} achieves the highest QED and the lowest Vina score, demonstrating its effectiveness in improving both drug-likeness and predicted binding affinity. Figure~\ref{fig:molecule} illustrates the scores of candidates from the final three \algshort{} iterations alongside an Auto-CoT candidate and three approved reference compounds.

\begin{table}[t]
\centering
\caption{\textbf{Comparison on molecule optimization.} Higher QED is better; lower Vina score is better.}
\label{tab:molecule_additional_baselines}
\resizebox{0.48\linewidth}{!}{%
\begin{tabular}{lccccc}
\toprule
Metric & Direct & Auto-CoT & USP & SR & UFPL \\
\midrule
QED $\uparrow$  & $0.55$ & $0.62$ & $0.59$ & $0.60$ & $\mathbf{0.72}$ \\
Vina $\downarrow$ & $-6.3$ & $-6.8$ & $-6.5$ & $-6.6$ & $\mathbf{-7.5}$ \\
\bottomrule
\end{tabular}
}
\end{table}

\subsection{Illustrative Example}
\label{sec:illustration}

In this section, we present prompts optimized by \algshort{} for the GSM8k and TruthfulQA datasets as an illustrations.

\begin{itemize}
\item \textbf{GSM8K (Llama-3.2-1B)}
Analyze the given examples to understand the problem structure and solution approach. Use these examples as a guide to reason through the current question step by step. Begin by identifying similar patterns or reasoning strategies in the examples, then apply them to the question at hand. If multiple approaches are possible, consider which one aligns best with the examples provided. Always ensure your reasoning is clear and logically follows from the given information and the examples. Once you have thoroughly analyzed the problem and considered the examples, provide your final answer.

\item \textbf{GSM8K (Qwen3-4B)}
Before answering the question, carefully rephrase it in your own words to ensure full understanding of all details, including units, conditions, and constraints. Use the provided in-context examples, which include questions, model-generated pseudo-labels, and answers, to inform your reasoning and align with expected task structure. Think step by step, clearly outlining your reasoning process and double-checking all calculations for accuracy. After completing your reasoning, verify that your final answer directly addresses the original question and is consistent with the steps taken. If the question is ambiguous or incomplete, indicate this and request clarification rather than making assumptions. This approach ensures alignment with reference answers, minimizes errors, and maintains attention to detail throughout the problem-solving process.

\item \textbf{TruthfulQA (Llama-3.2-1B)}
Use the in-context examples to inform your reasoning, but critically evaluate them for consistency, logic, and accuracy. When examples are ambiguous, incomplete, or potentially flawed, do not blindly follow them—instead, use your general knowledge to assess their validity. Balance adherence to the examples with sound reasoning and factual understanding, prioritizing accuracy over mere pattern matching. If there is ambiguity, consider multiple interpretations and explain your reasoning clearly. Identify the key elements of the question and apply logical analysis to arrive at a well-reasoned answer. Avoid relying solely on surface-level patterns (e.g., always choosing option C). Instead, focus on understanding the underlying logic and structure of the problem. Your response should demonstrate thoughtful analysis, even if the examples suggest a different approach. Always aim for clarity, correctness, and logical consistency in your reasoning.

\item \textbf{TruthfulQA (Qwen3-4B)}
Answer the following question by using the in-context examples as authoritative sources of factual information. When the examples provide clear factual answers—such as historical events, legal statutes, or scientific facts—prioritize those answers over logical, conceptual, or intuitive reasoning, even if they contradict common sense or expectations. Treat the examples as definitive guides to the correct factual conclusion, and align your response with the factual claims they present. When faced with ambiguity, focus on the specific factual patterns in the examples rather than general reasoning. Avoid relying on external knowledge or assumptions—strictly follow the information provided in the examples. After forming your answer, double-check it against the examples to ensure alignment, especially when the reasoning is complex or the answer appears counterintuitive. Your task is to prioritize factual accuracy based on the examples, even if it leads to an answer that seems less logically consistent.
\end{itemize}


\section{Prompt Implementation}
\label{sec:prompt-implementation}

In this section, we present the manual templates used by TextGrad for each dataset.

\subsection{Prompt Design in TextGrad}
\label{app:prompt_TextGrad}
For every task we compose a system prompt that fixes the global behavior of the task model and a task prompt that encodes the input variables.

The forward model receives the concatenation:
\texttt{<task-prompt> + <demonstrations> + <target example>}, where the task prompt is the system message, each demonstration is an earlier user/assistant turn with the model's reasoning and pseudo-label, and the target example is followed by a fixed answer-format line that is not part of the learnable prompt.

\textbf{Confidence filter.}
At initialization, data with at least four agreeing answers among five samples enter $H_0$ ($\gamma=0.65$). After each round, we update the pseudo-labels by majority voting over all answers collected so far. Confidence is the fraction of these votes supporting the current pseudo-label. We then select $H_{t+1}$ from the entire dataset using the same threshold. Both prompt training and demonstration selection use this updated set and its current pseudo-labels in the next round, allowing data to enter or leave the high-confidence set.

\textbf{Hyper-parameters.}
\begin{itemize}
  \item { Optimizer: \textsc{TGD} with Qwen3-14B (temperature 0.7 for the prompt rewrite, 0 for the critiques and feedback), mini-batches of 8 high-confidence questions;}
  \item { Prompt length: at most 180 words requested; rewrites longer than 270 words are rejected;}
  \item Demonstrations per example: $K=5$;
  \item \algshort{} iterations $T$: 3;
  \item { Update check: a fixed held-out set of $\min(32,\lfloor|H_0|/3\rfloor)$ questions sampled once from $H_0$ (none if $|H_0|<10$).}
\end{itemize}

\subsection{Prompt Design for Each Task}
\label{app:prompt_task}

We list the prompt design for each task in Table~\ref{tab:init-prompts}.

\begin{table*}[t]
\centering\small
\begin{tabular}{@{}>{\raggedright\arraybackslash}p{3.2cm} p{11cm}@{}}
\toprule
\textbf{Part} & \textbf{Text} \\
\midrule
Initial prompt $z_0$ (all tasks) & \texttt{Answer the following question. Think step by step before giving the final answer.} \\
Direct without CoT & \texttt{Answer the following question directly. Reply with the final answer line only, without explanation.} \\
Answer format, choice tasks & \texttt{Finish your response with a final line of the form 'Answer: $<$option$>$', where $<$option$>$ is one of: \{opts\}.} \\
Answer format, numeric tasks (GSM8K, SVAMP, ASDiv, MultiArith) & \texttt{Finish your response with a final line of the form 'Answer: $<$number$>$'.} \\
MMLU / HellaSwag / TruthfulQA / AQuA / CommonsenseQA / ARC-C / OpenBookQA input & \texttt{Question: \{question\} \textbackslash n Options: \textbackslash n A. \{option A\} \textbackslash n B. ...} \\
GSM8K / MultiArith input & \texttt{Question: \{question\}} \\
SVAMP / ASDiv input & \texttt{Question: \{body\} \{question\}} \\
StrategyQA input & \texttt{Question: \{question\} \textbackslash n Options: yes, no} \\
BBH input & \texttt{\{input\} (as released, options included)} \\
SST-2 input & \texttt{Review: \{sentence\} \textbackslash n Options: positive, negative} \\
RTE input & \texttt{Premise: \{sentence1\} \textbackslash n Hypothesis: \{sentence2\} \textbackslash n Options: entailment, not\_entailment} \\
\bottomrule
\end{tabular}
\caption{ Initial prompt, answer-format lines and input templates. The format line follows the target example and is not part of the learnable prompt.}
\label{tab:init-prompts}
\end{table*}

\section{Conclusion}
\label{sec:conclusion}
In this paper, we investigate specialization of general purpose LLM without retraining model parameters or relying on human supervision.
A straightforward approach is to generate ``high-confidence'' pseudo-supervised data and then apply in-context learning or prompt learning; however, using the few-shot examples only at inference rather than during prompt learning creates a mismatch between how the prompt is learned and how it is used.
We propose \algshort{}, an algorithm that learns prompts with demonstrations selected from a high-confidence set updated each round and refines predictions by pooling votes across rounds, thereby specializing general purpose LLMs for downstream tasks.
This joint optimization aligns prompt training with usage by requiring the learned prompt to produce consistent answers when pseudo-supervised data from the downstream task are used as few-shot examples.
In our binary model, demonstrations identify the label mapping within each group, and training selects the prompt that uses the predictive attribute across groups. Theoretical analysis bounds the resulting gain in expected accuracy over the initial pseudo-labels.
Experiments on benchmarks and a real-world molecule optimization task show the effectiveness of \algshort{}.

\newpage

\bibliography{./Refs_arxiv}

\newpage

\appendix

\end{document}